\pdfoutput=1
\documentclass{article}
\usepackage{fullpage}
\usepackage{times}
\usepackage{graphicx} 
\usepackage{caption,subcaption}
\usepackage{url}
\usepackage{epstopdf}
\usepackage{multirow}
\usepackage{amsmath,amsthm,amssymb}
\usepackage[title]{appendix}
       
\title{Convolutional Neural Networks for Text Categorization: \\ Shallow Word-level vs. Deep Character-level} 
\author{
Rie Johnson \\
RJ Research Consulting, NY, USA \\
{\tt riejohnson@gmail.com}\\
\and 
Tong Zhang \\
Rutgers University, NJ, USA \\
{\tt tzhang@stat.rutgers.edu}\\
}

\date{}

\begin{document} 
\maketitle

\begin{abstract} 
This paper reports the performances of shallow word-level convolutional neural networks (CNN), 
our earlier work (2015) \cite{JZ15a,JZ15b}, on the eight datasets with relatively large training data that were used for testing 
the very deep character-level CNN in Conneau et al. (2016) \cite{CSLB16}. 
Our findings are as follows. 
The shallow word-level CNNs achieve better error rates than the error rates reported in \cite{CSLB16} 
though the results should be interpreted with some consideration due to the unique pre-processing of \cite{CSLB16}. 
The shallow word-level CNN uses more parameters and therefore requires more storage than the deep character-level CNN; 
however, the shallow word-level CNN computes much faster.  
\end{abstract} 

\graphicspath{{figures/}} 

\newcommand{\cmt}[1]{{\em #1}\\}
\newcommand{\ZZL}{ZZL15}
\newcommand{\CSLByear}{Conneau et al. (2016)}
\newcommand{\CSLBcite}{\cite{CSLB16}}
\newcommand{\MDG}{MDG16}

\newcommand{\cnn}{CNN}
\newcommand{\cnns}{CNNs}
\newcommand{\wcnn}{word-CNN}
\newcommand{\ccnn}{char-CNN}
\newcommand{\wcnns}{word-CNNs}
\newcommand{\ccnns}{char-CNNs}
\newcommand{\Wcnn}{Word-CNN}
\newcommand{\Wcnns}{Word-CNNs}

\newcommand{\WW}{{\mathbf W}}
\newcommand{\xx}{{\mathbf x}}
\newcommand{\bb}{{\mathbf b}}

\section{Introduction}
\label{sec:intro}

Text categorization is the task of labeling documents, which has many important applications
such as sentiment analysis and topic categorization.  
Recently, several variations of 
{\em convolutional neural networks (\cnns)} \cite{LeCun+etal98} have been shown to achieve high accuracy 
on text categorization (see e.g., \cite{JZ15a,JZ15b,ZZL15,CSLB16} and references therein) 
in comparison with a number of methods including linear methods, which had long been the state of the art.  
{\em Long-Short Term Memory} networks (LSTMs) \cite{HS97} have also been shown to perform well on this task, 
rivaling or sometimes exceeding 
\cnns\ \cite{JZ16,MDG16}.  However, \cnns\ are particularly attractive since, 
due to their simplicity and parallel processing-friendly nature, 
training and testing of \cnns\ can be made much faster than LSTM to achieve similar accuracy \cite{JZ16},
and therefore \cnns\ have a potential to scale better to large training data.  
Here we focus on two \cnn\ studies that report high performances on categorizing long documents 
(as opposed to categorizing individual sentences): 
\begin{itemize}
\item
Our earlier work (2015) \cite{JZ15a,JZ15b}: shallow {\em word-level \cnns} (taking sequences of words as input), 
which we abbreviate as {\em \wcnn}. 
\item
\CSLByear\ \CSLBcite: very deep {\em character-level \cnns} (taking sequences of characters as input), 
which we abbreviate as {\em \ccnn}. 
\end{itemize}

Although both studies report higher accuracy than previous work on their respective datasets, 
it is not clear how they compare with each other due to lack of direct comparison. 
In \cite{CSLB16}, the very deep \ccnn\ was shown to perform well with larger training data (up to 2.6M documents) 
but perform relatively poorly with smaller training data; e.g., it underperformed linear methods 
when trained with 120K documents.  
In \cite{JZ15a,JZ15b} the shallow \wcnn\ was shown to perform well, using 
training sets (most intensively, 25K documents) that are mostly smaller than those used in \cite{CSLB16}. 
While these results imply that the shallow \wcnn\ is likely to outperform the deep \ccnn\ 
when trained with relatively small training sets such as those used in \cite{JZ15a,JZ15b}, 
the shallow \wcnn\ is untested on the training sets as large as those used in \CSLBcite. 
Hence, the purpose of this report is to fill the gap by testing the shallow \wcnns\ as in \cite{JZ15a,JZ15b} 
on the datasets 
used in \CSLBcite, for direct comparison with the results of very deep \ccnns\ reported in \CSLBcite. 

\paragraph{Limitation of work} 
In this work, our new experiments are limited to the shallow \wcnn\ as in \cite{JZ15a,JZ15b}.  
We do not provide new error rate results for the very deep \cnns\ proposed by \CSLBcite, 
and we only cite their results.  
Although 
it may be natural to assume 
that the error rates reported in \CSLBcite\ well represent the best performance 
that the deep \ccnns\ can achieve, 
we note that in \CSLBcite, documents were clipped and padded so that they all became 1014 characters long, 
and we do not know how this pre-processing affected their model accuracy.  
To experiment with \wcnn, 
we handle variable-sized documents as variable-sized as we see no merit in making them fixed-sized, 
though we reduce the size of vocabulary to reduce storage requirements.  
Considering that, we emphasize that 
this work is not intended to be a rigorous comparison of \wcnns\ and \ccnns; 
instead, it should be regarded as a report on the shallow \wcnn\ performance 
on the eight datasets used in \CSLBcite, 
referring to the results in \CSLBcite\ as the state-of-the-art performances.  

\subsection{Preliminary}

We start with briefly reviewing the very deep \wcnn\ of \cite{CSLB16} and the shallow \wcnn\ 
of \cite{JZ15a,JZ15b}.  
\subsubsection{Very deep character-level \cnns\ of \cite{CSLB16}} 

\cite{CSLB16} proposed very deep \ccnns\ and showed that 
their best performing models produced higher accuracy than 
their shallower models and previous deep \ccnns\ of \cite{ZZL15}.  
Their best architecture consisted of the following: 
\begin{itemize}
\item Character embedding of 16 dimensions. 
\item 29 convolution layers with the number of feature maps being 64, 128, 256, and 512. 
\item Two fully-connected layers with 2048 hidden units each, following the 29 convolution layers. 
\item One of the following three methods for downsampling to halve the temporal size: 
      setting stride to 2 in the convolution layer, $k$-max pooling, or max-pooling with stride 2. 
      Downsampling was done whenever the number of feature maps was doubled.  
\item $k$-max pooling with $k$=8 to produce 4096-dimensional input (per document) to the fully-connected layer.  
\item Batch normalization. 
\end{itemize}
The kernel size (`region size' in our wording) was set to 3 in every convolution layer. 
In addition, the results obtained by two more shallower architectures were reported.  
\cite{CSLB16} should be consulted for the exact architectures. 

\subsubsection{Shallow word-level \cnns\ as in \cite{JZ15a,JZ15b}}

\begin{figure}
\centering
%
\begin{subfigure}[b]{0.3\linewidth}
\begin{center}
\includegraphics[width=1.45in]{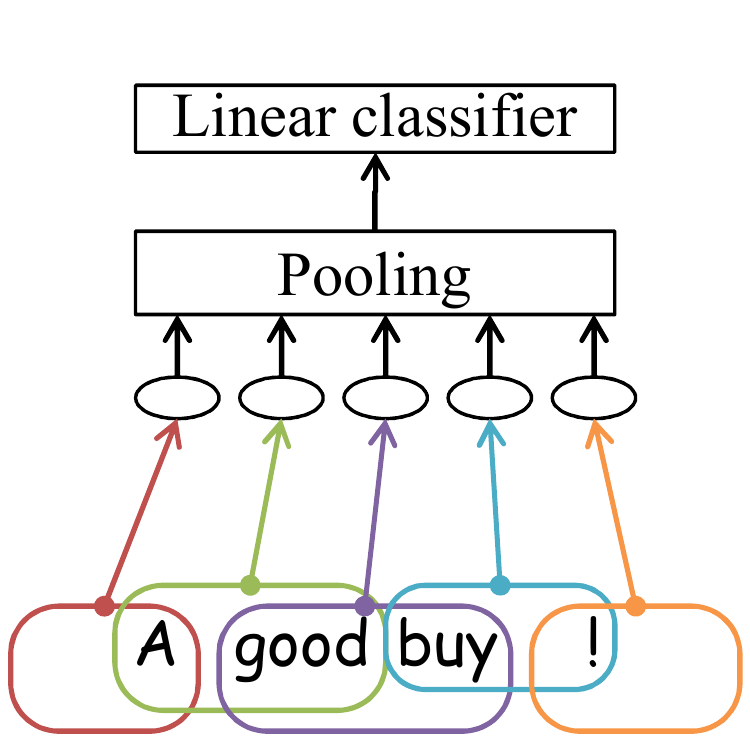}
\end{center}
\caption{\label{fig:onestep}
\wcnn\ (base model).  
}
\end{subfigure}%
%
\begin{subfigure}[b]{0.66\linewidth}
\begin{center}
\includegraphics[width=3.2in]{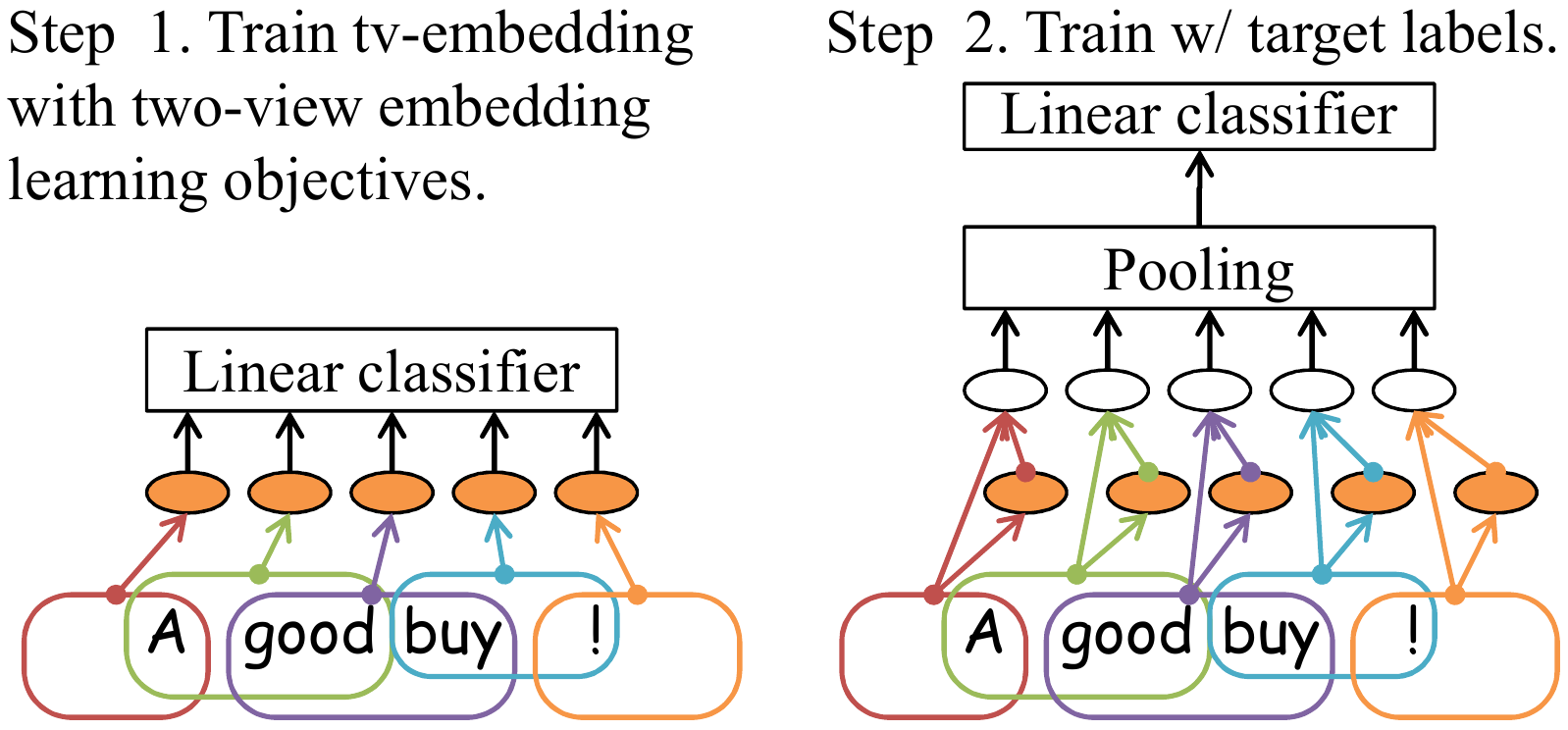}
\end{center}
\caption{\label{fig:twosteps} 
\wcnn\ with tv-embedding. 
}
\end{subfigure}
\caption{ \label{fig:wcnn} \small
  Shallow word-level \cnns.  In each oval, computation in the form of $\sigma(\WW \xx + \bb)$ takes place, 
  where $\xx$ is input, parameters $\WW$ and $\bb$ (shared within a layer) are trained, 
  and $\sigma$ is component-wise nonlinearity, typically $\sigma(x)=\max(0,x)$.  
  In the base model in (a), input $\xx$ is one-hot representation of each text region (e.g., ``good buy'').  
  In (b) we first train tv-embedding with two-view embedding learning objectives 
  and then use it to produce additional input to the base model.    
}
\end{figure}

Two types of \wcnn\ were proposed in \cite{JZ15a,JZ15b}, which are 
illustrated in Figure \ref{fig:wcnn}.  
One is a straightforward application of \cnn\ to text (the base model), and the other involves 
training of {\em tv-embedding} (`tv' stands for two views) to produce additional input to the base model. 
The models with tv-embedding produce higher accuracy provided that sufficiently large amounts of 
unlabeled data for tv-embedding learning are available.  
As discussed in \cite{JZ16}, the shallow \wcnn\ can be 
regarded
as a special case of a general 
framework which jointly trains a linear model with a non-linear
feature generator consisting of `text region embedding + pooling', where 
{\em text region embedding} is a loose term for 
a function that converts regions of text (word sequences such as ``good buy'') 
to vectors while preserving information relevant to the task of interest.  

\paragraph{\Wcnns\ without tv-embedding (base model)} 
In the simplest configuration of the shallow \wcnns, the region embedding is in the form of 
\[
  f(\xx) = \sigma\left( \WW \xx + \bb \right)
\] 
where $\sigma$ is a component-wise nonlinear function (typically $\sigma(x)=\max(x,0)$), 
input $\xx$ represents a text region via either the concatenation of one-hot vectors for the words in the region 
or the bow representation of the region,  and 
weight matrix $\WW$ and bias vector $\bb$ (shared within a layer) are trained.  
Note that  
when $\xx$ is the concatenation of one-hot vectors, 
$\WW \xx$ can be interpreted as summing {\em position-sensitive} word vectors, 
and when $\xx$ is the bow representation of the region, 
$\WW \xx$ can be interpreted as summing {\em position-insensitive} word vectors.  
Thus, in a sense, the region embedding $f(\xx)$ above {\em internally} and {\em implicitly} includes word embedding, 
as opposed to having an {\em external} and {\em explicit} word embedding layer before a convolution layer 
as in, e.g., \cite{Kim14}, 
which makes $\xx$ the concatenation of word vectors. 
See also the supplementary material of \cite{JZ15b} for the representation power analysis.  

As illustrated in Figure \ref{fig:wcnn} (a), 
$f(\xx)$ is applied to the text regions at every location of a document (ovals in the figure), 
and pooling aggregates the resulting region vectors into a document vector, 
which is used as features by a linear classifier.  

In our experiments with \wcnn\ without tv-embedding reported below, 
the one-hot representation used for $\xx$ was fixed to the concatenation of one-hot vectors with a vocabulary of the 30K most frequent words, 
and the dimensionality of region embedding (i.e., the number of feature maps) was fixed to 500.  
That is, our one-hot vectors were 30K-dimensional while any out-of-vocabulary word was converted to 
a zero vector, and the region embedding $f(\xx)$ produced 500-dimensional vectors for each region.  
Region size (the number of words in each region) was chosen from \{3,5\}.  
Based on our previous work, 
we performed max-pooling with $k$ pooling units (each of which covers $1/k$ of a document)
while setting $k=1$ on sentiment analysis datasets and choosing $k$ from $\{1,10\}$ on the others. 
The models described here also served as the base models of the \wcnn\ with tv-embedding described next. 

\paragraph{\Wcnns\ with tv-embedding} 
Training of \wcnns\ with tv-embedding is done in two steps, as shown in Figure \ref{fig:wcnn} (b) . 
First we train region {\em tv-embedding} (`tv' stands for two views) in the form of $f(\xx)$ above, 
with a two-view embedding learning objective 
such as `predict adjacent text regions (one view) based on a text region (the other view)'.  
This training can be done with unlabeled data.  
\cite{JZ15b} provides the definition and theoretical analysis of tv-embeddings.  
Next, we use the tv-embedding to produce additional input to the base model and train it with labeled data. 
This model can be easily extended to use multiple tv-embeddings, each of which, for example, uses 
a distinct vector representation of region, 
and so the region embedding function in the final model (hollow ovals in Figure \ref{fig:wcnn} (b)) can be written as: 
\[
  g(\xx,\{\xx^{(i)}\}_i) = \sigma\left(\WW \xx + \sum_i \WW^{(i)} \xx^{(i)} + \bb \right)~.  
\]
$\xx^{(i)}$ is the output of the tv-embedding indexed by $i$ applied to the corresponding text region. 
In \cite{JZ15b}, tv-embedding training was done using unlabeled data as an additional resource; 
therefore, the proposed models were semi-supervised models.  

In the experiments reported below,  
due to the lack of standard unlabeled data for the tested datasets,
we trained tv-embeddings on the labeled training data ignoring the labels; 
thus, the resulting models are supervised ones. 
We trained four tv-embeddings with four distinct one-hot representations of text regions 
(i.e., input to orange ovals in Figure \ref{fig:wcnn} (b)): 
bow representation with region size 5 or 9, and bag-of-\{1,2,3\}-gram representation with 
region size 5 or 9.   
To make bow representation for tv-embedding, we used a vocabulary of the 30K most frequent words, 
and to make the bag-of-\{1,2,3\}-gram representation, we used a vocabulary of the 200K most frequent \{1,2,3\}-grams. 
The dimensionality of tv-embeddings was 300 unless specified otherwise, and the 
dimensionality of $g(\cdot)$ was 500 (as in the base model); thus, we note that 
the dimensionality of internal vectors are comparable to those of the deep \ccnn\ of \CSLBcite, 
which are 64, 128, 256, and 512 as shown below. 
The rest of the setting was the same as the base model above.

\paragraph{Other two-step approaches}
Another two-step approach with \wcnns\ was studied by \cite{Kim14}, where the first step is pre-training of 
the word embedding layer (substituted by use of public word vectors in \cite{Kim14}), 
which is followed by a convolution layer.  
One potential advantage of our tv-embedding learning is that it can learn more complex information 
(embedding of {\em word sequences}) than word embedding (embedding of {\em single words in isolation}).  

\section{Experiments}

We report the experimental results of the shallow \wcnns\ 
in comparison with the results reported in \CSLBcite. 
The experiments can be reproduced using the code available at 
\url{riejohnson.com/cnn_download.html}. 

\subsection{Data and data preprocessing}

The eight datasets used in \CSLBcite\ are summarized in Table \ref{tab:cnnperf} (a).  
AG and Sogou are news, Dbpedia is an ontology, and Yelp and Amazon (abbreviated as `Ama') are reviews.  
`.p' (polarity) in the names of review datasets indicates that labels are either positive or negative, 
and `.f' (full) indicates that labels represent the number of stars.  
Yahoo contains questions and answers from the `Yahoo! Answers' website.  
On all datasets, classes are balanced.  Sogou consists of Romanized Chinese.  
The others are in English though some contain characters of other languages 
(e.g., Chinese, Korean) in small proportions.  
 
To experiment with the deep \ccnns, 
\CSLBcite\ converted upper-case letters to lower-case letters and used 72 characters 
(lower-case alphabets, digits, special characters, and special tokens for padding and 
out-of-vocabulary characters).  They {\em padded} the input text with a special token to a fixed size of 1014.  

To experiment with the shallow \wcnns, 
we also converted upper-case letters to lower-case letters. 
Unlike \CSLBcite, we handled variable-sized documents as variable-sized 
without any shortening or padding; however, we limited the vocabulary size 
to 30K words and 200K \{1,2,3\}-grams, 
as described above. 
To put it into perspective, 
the size of the complete word vocabulary of the largest training set (Ama.p) is 1.3M, 
and when limited to the words with frequency no less than 5, it is 221K.  
By comparison, a vocabulary of 30K sounds rather small, but it covers about 98\% of the text 
on Ama.p, and it appears to be sufficient for obtaining good accuracy.  

\subsection{Experimental details of word-level \cnns}

On all datasets, we held out 10K data points from the training set for use as validation data.  
Models were trained using the training set minus validation data, and model selection 
(or hyper parameter tuning) was done based on the performance on the validation data.  

Tv-embedding training was done as in \cite{JZ15b}; 
weighted square loss was minimized without regularization while the target regions (adjacent regions) were 
represented by bow vectors, and the data weights were set so that the negative sampling effect was achieved. 
Tv-embeddings were fixed (i.e., no weight updating) during the final training with labeled data.  

Training with labels (either with or without tv-embedding) was done as follows.  
A log loss (or cross entropy) with softmax was minimized.  
Optimization was done by mini-batch SGD with momentum 0.9 and the mini-batch size was set to 100.  
The number of epochs 
was fixed to 30 (except for AG, the smallest, for which it was fixed to 100), 
and the learning rate was reduced once by multiplying 0.1 
after 24 epochs (or 80 epochs on AG).  In all layers, weights were initialized by the Gaussian distribution of zero mean and 
standard deviation 0.01.  The initial learning rate was treated as a hyper parameter. 
Regularization was done by applying dropout with 0.5 to the input to the top layer and 
having a L2 regularization term with parameter 0.0001 on the top layer weights.  


\subsection{Performance results}

\begin{table}
\begin{center} \begin{tabular}{|l|c|c|c|c|c|c|c|c|c|}
\multicolumn{9}{l}{(a) Data statistics} \\
\hline
\multicolumn{2}{|l|}{}
                         &AG    & Sogou& Dbpedia & Yelp.p & Yelp.f & Yahoo & Ama.f & Ama.p \\
\hline                
\multicolumn{2}{|l|}{\# of training documents}
                         &120K & 450K & 560K    & 560K   & 650K   & 1.4M  &  3M  & 3.6M \\                    
\hline
\multicolumn{2}{|l|}{\# of test documents}
                         & 7.6K &  60K &  70K    &  38K   &  50K   &  60K  & 650K & 400K \\
\hline
\multicolumn{2}{|l|}{\# of classes}
                         &   4  &   5  &   14    &   2    &   5    &  10   &  5   & 2 \\
\hline
\multicolumn{2}{|l|}{Average length (words)}
                         & 45   &  578 &   55    & 153    &  155   & 112   &  93  & 91 \\ 
\hline
\multicolumn{2}{|l|}{Average length (characters)}
                         & 219  & 2709 &  298    &  710   &  718   & 519   & 441  & 432\\    
\hline

\multicolumn{9}{l}{} \\
\multicolumn{9}{l}{(b) Error rates (\%)} \\

\hline
Models & depth      &AG    & Sogou& Dbpedia & Yelp.p & Yelp.f & Yahoo & Ama.f & Ama.p \\
\hline  
Linear model {\em best} \cite{ZZL15} & 0 & 7.64 & 2.81 &    1.31  &   4.36   &  40.14 &    28.96 &    44.74  &    7.98 \\
\hline
\multirow{2}{*}{\ccnn\ {\em best} \cite{CSLB16}}
                                     & 9+2 &  9.17 & 3.58 & 1.35 & 4.88 & 36.73 & 27.60 & 37.95 & 4.70 \\
                                     & 29+2 & 8.67 & 3.18 & 1.29 & 4.28 & 35.28 & 26.57 &{\em 37.00}& 4.28 \\
\hline                                  
 \wcnn\ w/o tv-embed.         & 1 &\em{6.95}&\em{2.21}&\em{1.12}&\em{3.44}&\em{34.21}&\em{26.06}&     37.51 &\em{4.27}\\
 \wcnn\ w/ tv (300-dim)       & 2 &\bf{6.57}&\bf{1.89}&\bf{0.84}&\bf{2.90}&\bf{32.39}&\bf{24.85}& \bf{36.24}&\bf{3.79}\\ 
\hline
\end{tabular} \end{center}
\vskip -0.1in
\caption{ \label{tab:cnnperf}
  (a) Data statistics.  (b) Error rates (\%). \\
  {\small 
  `depth' counts the hidden layers with weights in the longest path.
  \cite{ZZL15} reported the results of several linear methods, and 
  we copied only the {\em best} results.  
  \cite{CSLB16} reported the results of deep \ccnn\ with three downsampling methods, and 
  we copied only the {\em best} results.  
  The \wcnn\ results are our new results. 
  The best (or second best) results are shown in bold (or italic) font, respectively.}
}
\end{table}

\paragraph{Error rates} 
In Table \ref{tab:cnnperf} (b), we show the error rate results of the shallow \wcnn\ 
in comparison with the best results of the deep \ccnn\ reported in \CSLBcite\ and 
the best results of linear models reported in \cite{ZZL15}.  
On each dataset, the best results are shown in bold and the second best results are 
shown in the italic font.  

On all datasets, the shallow \wcnn\ with tv-embeddings performs the best.  
The second best performer is the shallow \wcnn\ without tv-embedding on all but Ama.f (Amazon full).  
Whereas the deep \ccnn\ underperforms traditional linear models when 
training data is relatively small, 
the shallow \wcnns\ with and without tv-embedding clearly outperform them on all the datasets. 
We observe that, as in our previous work \cite{JZ15b}, 
additional input produced by tv-embeddings led to substantial improvements. 

The performances of \wcnn\ without tv-embedding might be further improved by having 
multiple region sizes \cite{JZ15a,Kim14}, but for simplicity, we did not attempt it in this work.  



\newcommand{\tworow}[1]{\multirow{2}{*}{#1}}
\begin{table}
\begin{center} \begin{tabular}{|ll|c|l|r|r|c|}
\hline
      && \tworow{Depth} &\multicolumn{1}{|c|}{Dimensionality of}       &\tworow{\#param} &\tworow{Time} & Error\\
      &&                &\multicolumn{1}{|c|}{layer outputs (\#layers)} &                 &             & rate(\%)\\
\hline                        
\multirow{4}{*}{\ccnn\ \CSLBcite}
                     && \tworow{9+2}  
                             & 16(1), 64(3), 128(2), & \tworow{2.2M} & \tworow{$\dagger$215} &\tworow{36.73}\\
                     &&      & 256(2), 512(2), 2048(2) &             &                       &\\
\cline{3-7}                     
                     && \tworow{29+2}  
                             & 16(1), 64(11), 128(10), & \tworow{4.6M} & \tworow{$\ddagger$700} &\tworow{35.28}\\
                     &&      & 256(4), 512(4), 2048(2) &               &                       &\\            
\hline
\multirow{3}{*}{\wcnn} 
               &\multicolumn{1}{|l|}{w/o tv-embed.}&   1   & 500(1)         &  45M &  6 & 34.21 \\
\cline{2-7}
               &\multicolumn{1}{|l|}{w/ 2 tv (100-dim)}&   2   & 100(2), 500(1) &  68M & 21 & 32.77\\               
\cline{2-7}
               &\multicolumn{1}{|l|}{w/ 4 tv (100-dim)}&   2   & 100(4), 500(1) &  91M & 36 & 32.55\\
\cline{2-7}
               &\multicolumn{1}{|l|}{w/ 4 tv (300-dim)}&   2   & 300(4), 500(1) & 184M & 72 & 32.39\\
\hline
\end{tabular} \end{center}
\vskip -0.1in
\caption{ \label{tab:sizetime}
Model size and computation time. 
{\small 
`Time': Elapsed time (seconds) for testing on the Yelp.f test data using Tesla M2070.  
It excludes preprocessing for input vector generation including 
one-hot vector manipulation (concatenation/bow generation) for \wcnn. 
Error rates are also on Yelp.f.  
The shallow \wcnn\ has more parameters but computes faster than the deep \ccnn.  
Information on the deep \ccnn\ is from \CSLBcite\ except for `Time'.  
\\
$\dagger$ $\ddagger$ 
Processing time depends on implementation, and 
test time of the deep \ccnn\ was measured using our implementation. 
As described in \CSLBcite, we clipped and padded documents so that the documents all became 1014 characters long. 
}
}
\end{table}

\begin{table}
\begin{center} \begin{tabular}{|l|c|c|c|c|c|c|c|c|c|}
\hline
          &AG    & Sogou& Dbpedia & Yelp.p & Yelp.f & Yahoo & Ama.f & Ama.p \\
\hline          
\wcnn\ w/ 4 tv (100-dim) & 6.57 & 1.96 & 0.84    & 2.97   & 32.55  & 25.14 & 36.52 & 3.90 \\
\hline
\wcnn\ w/ 4 tv (300-dim) & 6.57 & 1.89 & 0.84    & 2.90   & 32.39  & 24.85 & 36.24 & 3.79 \\ 
\hline
\end{tabular} \end{center}
\caption{ \label{tab:smalldim}
Error rates of the shallow \wcnn\ with tv-embeddings of 100 dimensions (`w/ 4 tv(100-dim').  \\
{\small 
  `w/ 4 tv (300-dim)' was copied from Table \ref{tab:cnnperf} (b) for easy comparison.  
}
}
\end{table}

\paragraph{Model size and computation time} 
In Table \ref{tab:sizetime}, we observe that, 
{\em compared with the deep \ccnn, the shallow \wcnn\ has more parameters but computes much faster}.  
Although the table shows computation time and error rates on one particular dataset (Yelp.f), 
the observation was the same on the other datasets.  
The shallow \wcnn\ has more parameters 
because the number of parameters mostly depends on the vocabulary size, which is large 
with \wcnn\ (30K and 200K in our experiments) and small with \ccnn\ (72 in \CSLBcite). 
Nevertheless, computation of the shallow \wcnn\ can be made much faster
than the deep \ccnn\ for three reasons\footnote{
  Computation time depends on the specifics of both implementation and environment including hardware.
  Our discussion here assumes parallel processing and efficient handling of sparse matrices (whose components are mostly zero), 
  and otherwise it is general. 
}. 
First, with implementation to handle sparse data efficiently, 
computation of shallow \wcnn\ does {\em not} depend on the vocabulary size. 
For example, when $\xx$ is the concatenation of $p$ one-hot vectors of dimensionality $v$ (vocabulary size), 
computation time of $\WW \xx$ (the most time-consuming step) depends {\em not} on $v$ (e.g., 30K) 
but on $p$ (e.g., 3) since we only need to multiply nonzero elements of $\xx$ with the weights in $\WW$.  
Second, character-based methods need to process about five times more text units than word-based methods; 
compare the rows of average length in words and characters in Table \ref{tab:cnnperf} (b).  
Third, a deeper network is less parallel processing-friendly 
since many layers have to be processed sequentially.   

If we reduce the dimensionality of tv-embedding from 300 to 100, 
the number of parameters can be reduced to a half with a small degradation of accuracy, 
as shown in Table \ref{tab:sizetime}; 
more error rate results with 100-dim tv-embedding are shown in Table \ref{tab:smalldim}.  
Reducing the number of tv-embeddings from four to two also reduces the number of parameters 
with a small degradation of accuracy (`w/ 2 tv (100-dim)' in Table \ref{tab:sizetime}). 

\paragraph{Summary of the results}
\begin{itemize}
\item 
The shallow \wcnns\ as in \cite{JZ15a,JZ15b} generally achieved better error rates 
than those of the very deep \ccnns\ reported in \CSLBcite.  
\item
The shallow \wcnn\ computes much faster than the very deep \ccnn. 
This is because the deep \ccnn\ needs to process more text units as there are many more characters than 
words per document, and because many layers need to be processed sequentially.  
This is a practical advantage of the shallow \wcnn.  
\item
The shallow \wcnns\ use more parameters and therefore require more storage, 
which is a drawback in storage-tight situations. 
Reducing the number and/or dimensionality of tv-embeddings reduces the number of parameters 
though it comes with the expense of a small degradation of accuracy.  
\end{itemize}

\bibliography{cnn-perf}
\bibliographystyle{plain}

\end{document}